\documentclass[sigconf]{acmart}
\usepackage{subcaption}
\usepackage{enumitem}

\AtBeginDocument{%
  }

\copyrightyear{2026}
\acmYear{2026}
\setcopyright{cc}
\setcctype{by}
\acmConference[WSDM Companion '26]{The Nineteenth ACM International Conference on Web Search and Data Mining}{February 22--26, 2026}{Boise, ID, USA}
\acmBooktitle{The Nineteenth ACM International Conference on Web Search and Data Mining (WSDM Companion '26), February 22--26, 2026, Boise, ID, USA}
\acmPrice{}
\acmDOI{10.1145/3779211.3793174}
\acmISBN{979-8-4007-2358-2/2026/02}

\begin{document}

\title{WISE: Web Information Satire and Fakeness Evaluation}

\author{Gaurab Chhetri}
\affiliation{%
  \institution{Texas State University}
  \city{San Marcos, Texas}
  \country{USA}}
\email{gaurab@txstate.edu}
\orcid{0009-0000-0124-4814}

\author{Subasish Das}
\affiliation{%
  \institution{Texas State University}
  \city{San Marcos, Texas}
  \country{USA}}
\email{subasish@txstate.edu}
\orcid{0000-0002-1671-2753}

\author{Tausif Islam Chowdhury}
\affiliation{%
  \institution{Texas State University}
  \city{San Marcos, Texas}
  \country{USA}}
\email{sgp98@txstate.edu}
\orcid{0009-0008-2385-8719}

\renewcommand{\shortauthors}{Chhetri et al.}

\begin{abstract}
Distinguishing fake or untrue news from satire or humor poses a unique challenge due to their overlapping linguistic features and divergent intent. This study develops WISE (Web Information Satire and Fakeness Evaluation) framework, which benchmarks eight lightweight transformer models alongside two baseline models on a balanced dataset of 20,000 samples from Fakeddit, annotated as either fake news or satire. Using stratified 5-fold cross-validation, we evaluate models across comprehensive metrics including accuracy, precision, recall, F1-score, ROC-AUC, PR-AUC, MCC, Brier score, and Expected Calibration Error. Our evaluation reveals that MiniLM, a lightweight model, achieves the highest accuracy (87.58\%) among all models, while RoBERTa-base achieves the highest ROC-AUC (95.42\%) and strong accuracy (87.36\%). DistilBERT offers an excellent efficiency-accuracy trade-off with 86.28\% accuracy and 93.90\% ROC-AUC. Statistical tests confirm significant performance differences between models, with paired t-tests and McNemar tests providing rigorous comparisons. Our findings highlight that lightweight models can match or exceed baseline performance, offering actionable insights for deploying misinformation detection systems in real-world, resource-constrained settings.
\end{abstract}



\received{27 November 2025}

\maketitle

\begin{figure}[H]
    \centering
    \includegraphics[width=\linewidth]{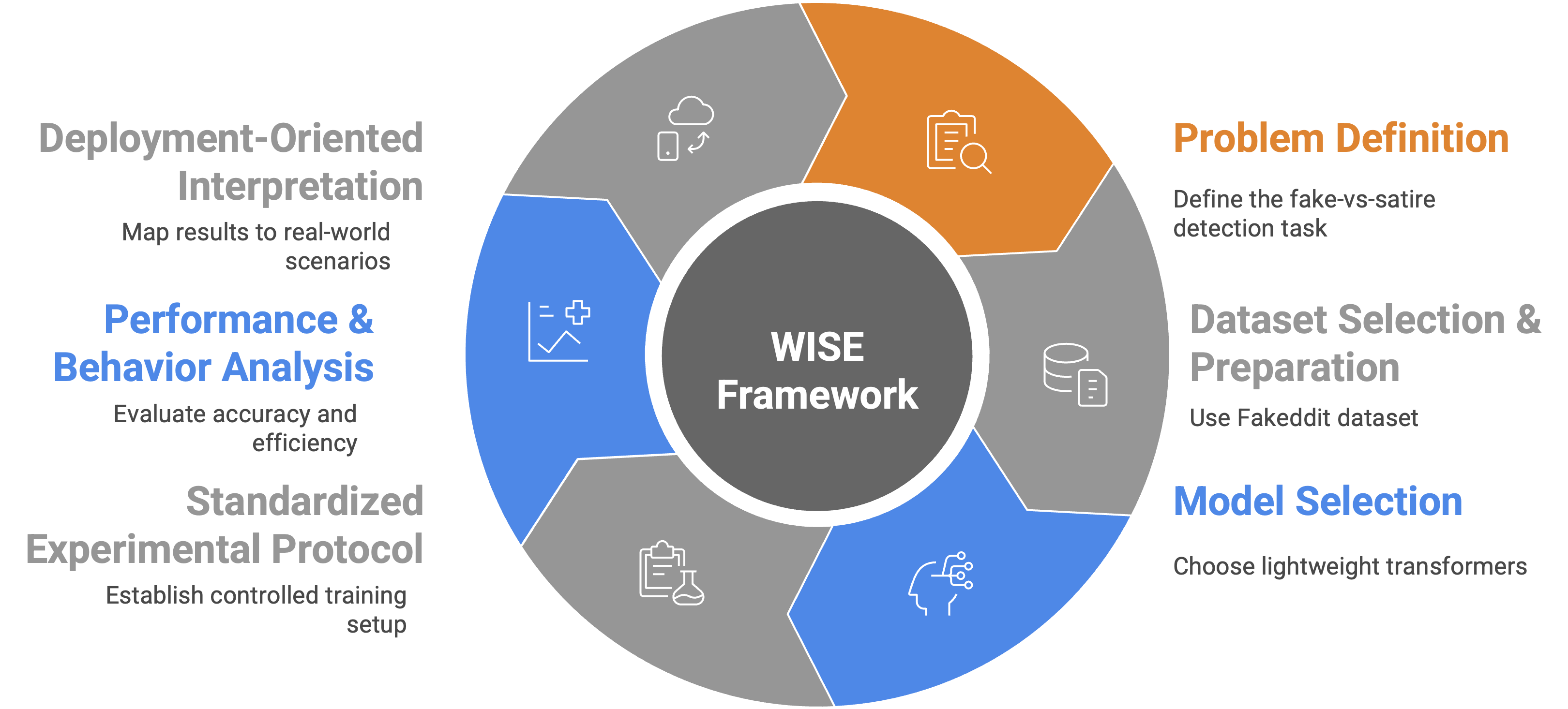}
    \caption{Overview of the WISE framework}
    \label{fig:wise-ov}
\end{figure}

\section{Introduction}
\label{sec:intro}

The rapid proliferation of online information has intensified the need to distinguish malicious fake news from satirical content. Fake news, defined as intentionally false information presented as factual, has become a major societal concern, often conflated with other misleading content such as satire and factual errors \cite{golbeck2018fake}. False news spreads faster and farther than truthful news, largely driven by human behavior rather than automation \cite{vosoughi2018spread,aral2018wired}, and can cause serious harm, as seen in conspiracy-driven vandalism incidents. Addressing misinformation requires avoiding the conflation of fake news with satire, as the latter mimics news style while differing in intent. Classifiers trained to separate satire from real news often also detect fake news, reflecting overlapping linguistic patterns \cite{horne2017fake_satire_similarity}, yet satire aims for humor and critique rather than deception. Mislabeling satire or emotion associated with it undermine its role in public discourse, where irony and exaggeration are used for social commentary \citep{das_city_2023, das_artificial_2023}. Thus, automated systems must reliably separate harmful misinformation from benign satire to protect both information integrity and satirical expression.

Fake news detection has evolved from feature-based classifiers using linguistic and stylistic cues (e.g., n-grams, readability, sentiment) to deep learning methods. Early approaches with support vector machines (SVMs) or Naive Bayes on handcrafted features achieved moderate success \cite{rastogi2023fake_news_survey}. Convolutional Neural Networks (CNNs) and Recurrent Neural Networks (RNNs) improved representation learning, and transformer-based models like Bidirectional Encoder Representations from Transformers (BERT) further advanced accuracy via contextualized understanding \cite{kaliyar2021fakebert}. However, most research frames detection as real vs. fake or as part of broader multi-class misinformation taxonomies, overlooking the fake-vs-satire challenge. Satirical writing employs rhetorical devices such as irony and sarcasm, complicating detection. Only a small body of work, such as Levi et al. \cite{levi2019fake_vs_satire}, addresses this distinction using semantic, linguistic, and coherence features with contextual models. Multimodal satire detection incorporating visual cues shows potential \cite{nguyen2021multimodal}, but there remains limited evaluation of lightweight transformers on fake-vs-satire tasks. This gap limits understanding of whether compact models can capture these nuances while remaining efficient—an essential consideration for real-world deployments.

In this study, we benchmark eight lightweight transformer models alongside two baseline models for distinguishing fake news from satire. We use a balanced subset of the Fakeddit dataset \cite{nakamura2019fakeddit}, comprising 20,000 samples (10,000 satire and 10,000 fake news), ensuring robust evaluation through stratified 5-fold cross-validation. We assess comprehensive classification metrics including accuracy, precision, recall, F1-score (macro and per-class), Receiver Operating Characteristic Area Under the Curve (ROC-AUC), Precision-Recall Area Under the Curve (PR-AUC), Matthews Correlation Coefficient (MCC), Brier score, and Expected Calibration Error (ECE), alongside training and inference efficiency. Our contributions are: (1) the first systematic benchmark of multiple lightweight transformers on a large-scale fake-vs-satire dataset using rigorous cross-validation, (2) comprehensive evaluation with full statistical testing including paired t-tests, McNemar tests, and DeLong tests, (3) analysis of how depth, parameter sharing, and distillation influence detection of satirical vs. deceptive cues, and (4) deployment-oriented insights for building high-accuracy, resource-efficient misinformation detection systems.

\section{Related Works}
\label{sec:relatedworks}

\subsection{Fake News Detection}

Fake news detection techniques have progressed markedly over the past decade. Early approaches relied on handcrafted linguistic features, for example, n-gram frequency patterns, lexical complexity, and sentiment cues, to distinguish deceptive content \cite{potthast2018stylometric}. Such methods showed that deceptive writing often exhibits telltale stylistic signs, such as negative affect or unusual phrasing, that can be captured via bag-of-words or parse-tree representations \cite{rubin2016fake}. As the field advanced, machine learning and deep neural networks gained prominence. Researchers began applying convolutional and recurrent neural networks to automatically learn features from text, yielding better accuracy than earlier feature-engineered models \cite{zhang2020overview}. For instance, long short-term memory and convolutional architectures, sometimes combined with metadata, outperformed traditional classifiers on benchmark fake news datasets \cite{ruchansky2017csi}. The emergence of transformer-based language models like BERT has since revolutionized NLP, and several studies report that fine-tuning such models significantly boosts fake news classification performance \cite{kaliyar2021fakebert,jawahar2020bertstructure}. BERT-based detectors, such as FakeBERT, can capture subtle linguistic nuances and contextual cues, leading to state-of-the-art results in identifying misinformation \cite{kaliyar2021fakebert}. However, most of this literature frames fake news detection as a binary classification, distinguishing only between real and fake news \cite{rubin2015deception}. Relatively little work has addressed finer-grained distinctions between different kinds of misleading content. Notably, Rubin \cite{rubin2015deception} identified categories like outright fabrications, large-scale hoaxes, and satire as separate challenges, highlighting that current systems typically do not attempt to differentiate satire from intentional fake news. This gap motivates specialized investigations into sub-types of misinformation.

\subsection{Satire Detection}
In contrast to general fake news detection, the automatic recognition of satirical news has received comparatively less scholarly attention. Satire is a form of deceptive news that uses humor, irony, and exaggeration not to mislead maliciously, but to entertain or critique \cite{bourgonje2017clickbait, mihalcea2006humor}. These traits make satire especially challenging for algorithms to identify, as the language often mimics legitimate news while embedding subtle cues such as absurd scenarios or sarcastic tone that signal its true intent \cite{burfoot2009satire}. Early studies approached satire detection through linguistic and stylistic features. For example, Burfoot and Baldwin \cite{burfoot2009satire} and later Rubin et al. \cite{rubin2016satire_detection} used features like absurdity, slang, negative affect, and punctuation patterns combined with support vector machine classifiers to successfully distinguish satirical articles. Rubin’s system achieved an F-score around 0.87 by leveraging such handcrafted cues \cite{rubin2016satire_detection}. Some subsequent works experimented with deep learning. Yang et al. \cite{yang2017humor} proposed a hierarchical attention network to capture satire at both paragraph and article levels. However, the application of advanced neural architectures to satire-specific detection remains limited. Only a few datasets exist for this task; one prominent example is the Golbeck et al. \cite{golbeck2018fake} dataset of online news labeled as fake or satire. While this resource has enabled initial studies on satirical news classification, the use of transformer-based architectures such as BERT or its variants for satire detection is still in its early stages. Research has only begun to explore whether these models can reliably capture the nuanced humor and irony in satire, and current findings suggest it remains an open problem requiring further investigation~\cite{seth2024bert}.

\subsection{Lightweight Transformer Models}
The powerful performance of large transformers such as BERT comes at the cost of high model complexity and substantial computational resource requirements. This challenge has motivated the development of lightweight transformer models that preserve strong language understanding capabilities while operating more efficiently. One prominent approach is knowledge distillation, as demonstrated by TinyBERT, which compresses BERT into a smaller model by transferring knowledge from a large teacher model to a compact student model \cite{jiao2020tinybert}. A four-layer TinyBERT can achieve approximately 96–97\% of its teacher’s accuracy on General Language Understanding Evaluation (GLUE) benchmarks while being more than seven times smaller and nearly nine times faster during inference \cite{jiao2020tinybert}. Another strategy is exemplified by ALBERT (A Lite BERT), which employs parameter-reduction techniques such as factorized embedding layers and cross-layer parameter sharing to drastically reduce the number of parameters without significantly impacting performance \cite{lan2020albert}. ALBERT achieved new state-of-the-art results on benchmarks like Stanford Question Answering Dataset (SQuAD) and GLUE with 18 times fewer parameters than BERT-large. Similarly, MiniLM uses a deep self-attention distillation technique focusing on key transformer attention components, enabling the student model to retain over 99\% of BERT’s performance on tasks such as question answering and natural language inference while using only about half the model size and computation \cite{wang2019glue}. Additional compact models include DistilBERT, which is approximately 40 percemt smaller and 60\% faster than BERT-base while maintaining around 95–97\% of its language understanding capability \cite{sanh2019distilbert}, and MobileBERT, which optimizes BERT for mobile devices through a bottleneck architecture \cite{sun2020mobilebert}. These lightweight transformers have demonstrated strong performance across diverse NLP tasks, showing that substantial model compression is possible with minimal accuracy loss, making them attractive for deployment in resource-constrained or real-time fake news detection systems.

\subsection{Benchmark Studies}
There have been numerous benchmark studies evaluating transformer models on general NLP tasks, yet none have specifically addressed the challenge of distinguishing fake news from satire. General benchmark frameworks such as GLUE and SuperGLUE have facilitated systematic comparisons of models including BERT, RoBERTa, and ALBERT on tasks such as text classification, entailment, and question answering \cite{wang2019glue, wang2020minilm}. These evaluations have driven progress in the field by identifying the architectures that perform best under various conditions. However, the satire-versus-fake news classification problem has largely been absent from these benchmark suites. A notable exception is the dataset introduced by Golbeck et al. \cite{golbeck2018fake}, which contains 283 fake news articles and 203 satire articles within the same political domain, each with verified labels. This dataset demonstrated strong inter-annotator agreement (approx.= 0.69) and 84\% labeling accuracy, making it a reliable resource for satire detection research. It has been used in several studies to test classification models, with reported baseline accuracies generally in the mid-70\% to low-80\% range for distinguishing satire from fake news \cite{das2020heuristic}. Another dataset, FakeNewsNet and its derivative Fakeddit, includes satire as one of multiple misinformation categories \cite{shu2020fakenewsnet,nakov2021clef}. However, these datasets are not exclusively focused on satire-vs-fake classification and often include social media context that differs from mainstream news \citep{das_mapping_2024}. Overall, there is a clear lack of comprehensive benchmarks dedicated to this specific classification task. The present study addresses this gap by systematically evaluating multiple lightweight transformer models on the satire-versus-fake news distinction, providing the first in-depth comparison of their performance in this specialized misinformation detection domain.

\section{Methodology}
\label{sec:method}
This section outlines the dataset, models, and experimental procedures used in our study, with the complete workflow summarized in Figure \ref{fig:wise-ov}.

\subsection{Dataset}
\label{sec:dataset}

We used a balanced subset of the Fakeddit dataset \cite{nakamura2019fakeddit}, a large-scale multimodal dataset for fake news detection collected from Reddit. Fakeddit contains posts labeled across six categories: True, Satire/Parody, Misleading Content, Manipulated Content, False Connection, and Imposter Content. For our binary classification task, we curated a balanced dataset focusing on the distinction between satire and fake news.

\subsubsection{Dataset Preparation}
We extracted samples from the Fakeddit training, validation, and test splits, selecting posts with clean titles longer than 50 characters to ensure sufficient content for classification. From the available labels, we selected:
\begin{itemize}[leftmargin=*]
    \item \textbf{Satire}: 10,000 samples labeled as ``Satire/Parody'' (label 1)
    \item \textbf{Fake News}: 10,000 samples combining ``Misleading Content'' (label 2) and ``Manipulated Content'' (label 3), with 5,000 samples from each category
\end{itemize}

This selection strategy ensures a balanced dataset while representing different types of misinformation. The ``Misleading Content'' and ``Manipulated Content'' categories both represent intentionally deceptive content, making them appropriate proxies for fake news in contrast to satirical content. The final dataset contains 20,000 samples with perfect class balance (see~\autoref{tab:dataset_stats}).

\begin{table}[h]
\centering
\small
\caption{Dataset Statistics}
\label{tab:dataset_stats}
\begin{tabular}{lccc}
\toprule
\textbf{Category} & \textbf{Count} & \textbf{Percentage} & \textbf{Avg. Length} \\
\midrule
Satire & 10,000 & 50.0\% & 82.4 \\
Fake News & 10,000 & 50.0\% & 83.1 \\
\midrule
\textbf{Total} & \textbf{20,000} & \textbf{100\%} & \textbf{82.7} \\
\bottomrule
\end{tabular}
\end{table}

The dataset exhibits moderate text length variation, with titles ranging from 51 to 296 characters. The mean text length is 82.7 characters, indicating that most samples are relatively concise Reddit post titles. This length distribution is well-suited for transformer-based classification, as most samples fit comfortably within standard token limits without requiring truncation. The balanced class distribution eliminates potential bias from class imbalance, allowing for fair evaluation across all metrics.

\subsection{Models}
We benchmarked eight lightweight transformer encoders alongside two baseline models, chosen for their balance of predictive performance and computational efficiency, which makes them suitable for misinformation detection in resource-constrained environments. Each model is publicly available via the Hugging Face model hub for reproducibility. Table~\ref{tab:model_specs} provides detailed specifications for each model.

\begin{table*}[h]
\centering
\small
\caption{Model specifications aligned with actual Hugging Face checkpoints used}
\label{tab:model_specs}
\begin{tabular}{lcccc}
\toprule
\textbf{Model (Checkpoint)} & \textbf{Layers} & \textbf{Hidden Size} & \textbf{Attention Heads} & \textbf{Parameters (approx.)} \\
\midrule
\multicolumn{5}{l}{\textit{Lightweight Models}} \\
TinyBERT (\texttt{prajjwal1/bert-tiny}) & 2 & 128 & 2 & 4.39M \\
TinyBERT4L (\texttt{huawei-noah/TinyBERT\_General\_4L\_312D}) & 4 & 312 & 12 & 14.5M \\
ALBERT (base-v2) (\texttt{albert/albert-base-v2}) & 12 & 768 & 12 & 12M \\
MiniBERT (\texttt{prajjwal1/bert-mini}) & 4 & 256 & 4 & $\sim$8M \\
MiniLM (\texttt{microsoft/MiniLM-L12-H384-uncased}) & 12 & 384 & 12 & 33M \\
Small-BERT-L2 (\texttt{google/bert\_uncased\_L-2\_H-256\_A-4}) & 2 & 256 & 4 & $\sim$4M \\
DistilBERT (\texttt{distilbert-base-uncased}) & 6 & 768 & 12 & 66M \\
ELECTRA-small (\texttt{google/electra-small-discriminator}) & 12 & 256 & 4 & 14M \\
\midrule
\multicolumn{5}{l}{\textit{Baseline Models}} \\
BERT-base (\texttt{bert-base-uncased}) & 12 & 768 & 12 & 110M \\
RoBERTa-base (\texttt{roberta-base}) & 12 & 768 & 12 & 125M \\
\bottomrule
\end{tabular}
\end{table*}

\subsubsection{Lightweight Models}
\begin{enumerate}[leftmargin=*]
    \item \textbf{TinyBERT}\footnote{\url{https://huggingface.co/prajjwal1/bert-tiny}} is a highly compact distilled variant of BERT with 2 transformer layers and a hidden size of 128. Its design prioritizes fast inference and minimal memory usage while preserving core language understanding capabilities.
    
    \item \textbf{TinyBERT4L}\footnote{\url{https://huggingface.co/huawei-noah/TinyBERT_General_4L_312D}} is a 4-layer variant with 312 hidden dimensions and 12 attention heads, designed as a general-purpose version fine-tuned across a variety of NLP tasks.
    
    \item \textbf{ALBERT (base v2)}\footnote{\url{https://huggingface.co/albert/albert-base-v2}} employs parameter sharing across layers and factorized embedding parameterization to drastically reduce the number of trainable parameters without sacrificing model depth.
    
    \item \textbf{MiniBERT}\footnote{\url{https://huggingface.co/prajjwal1/bert-mini}} is a compact BERT variant with 4 transformer layers and a 256-dimensional hidden size, designed for constrained deployments.
    
    \item \textbf{MiniLM}\footnote{\url{https://huggingface.co/microsoft/MiniLM-L12-H384-uncased}} uses a knowledge distillation strategy that transfers self-attention value relation knowledge from large teacher models into a student network with 12 layers and a hidden size of 384.
    
    \item \textbf{Small-BERT-L2}\footnote{\url{https://huggingface.co/google/bert_uncased_L-2_H-256_A-4}} is a minimal BERT architecture with 2 layers, 256 hidden dimensions, and 4 attention heads, representing an extreme compression approach.
    
    \item \textbf{DistilBERT}\footnote{\url{https://huggingface.co/distilbert-base-uncased}} is a distilled version of BERT that is 40\% smaller and 60\% faster while retaining 97\% of BERT's performance, using knowledge distillation during pre-training.
    
    \item \textbf{ELECTRA-small}\footnote{\url{https://huggingface.co/google/electra-small-discriminator}} uses a replaced token detection pre-training task instead of masked language modeling, enabling more efficient learning with smaller models.
\end{enumerate}

\subsubsection{Baseline Models}
\begin{enumerate}[leftmargin=*]
    \item \textbf{BERT-base}\footnote{\url{https://huggingface.co/bert-base-uncased}} serves as a standard baseline with 12 layers, 768 hidden dimensions, and 110M parameters, representing the original transformer architecture.
    
    \item \textbf{RoBERTa-base}\footnote{\url{https://huggingface.co/roberta-base}} is an optimized version of BERT with improved training procedures, serving as a strong baseline for comparison.
\end{enumerate}

These architectures embody different compression and optimization strategies, ranging from aggressive parameter reduction to sophisticated distillation. Evaluating them side-by-side on the fake-news–versus–satire classification task provides insight into how such design choices affect both classification accuracy and computational efficiency.

\subsection{Experimental Design and Training Methodology}

We benchmark eight lightweight transformer models (TinyBERT, TinyBERT4L, ALBERT, MiniBERT, MiniLM, Small-BERT-L2, DistilBERT, and ELECTRA-small) alongside two baseline models (BERT-base and RoBERTa-base) on distinguishing fake news from satire, using stratified 5-fold cross-validation for robust evaluation. The study pursues three goals: establish comprehensive baselines for each architecture, quantify accuracy versus efficiency trade-offs, and surface deployment choices for different application needs. This multi-angle evaluation demonstrates how well lightweight transformers work in practical misinformation detection settings.

\subsubsection{Data Splitting and Cross-Validation}
We employ stratified 5-fold cross-validation to ensure robust evaluation and reduce variance in performance estimates. The dataset of 20,000 samples is split into 5 folds using StratifiedKFold while maintaining class balance across all splits. For each fold, the training-validation set (80\% of data) is further split into training (70\%) and validation (15\%) sets, with the remaining 15\% held out as the test set. This nested splitting approach ensures that each fold has independent train, validation, and test sets while maintaining class balance. This approach provides multiple independent evaluations per model, enabling statistical significance testing and more reliable performance estimates. Each model is evaluated across all 5 folds, resulting in 5 independent performance measurements per model.

\subsubsection{Preprocessing and Tokenization}
Preprocessing preserves linguistic integrity while preparing inputs for each model's tokenizer. Text samples are normalized with consistent whitespace and special character handling, then tokenized with each model's native tokenizer from the Hugging Face Transformers library \cite{wolf2020transformers}. Inputs are capped at 256 tokens (configurable via \texttt{max\_length}), which comfortably covers the dataset's text length distribution (mean 82.7 characters, range 51-296). This token limit ensures efficient processing while maintaining sufficient context for classification. Tokenized inputs are cached to disk to avoid redundant tokenization across folds and models, improving experimental efficiency.

\subsubsection{Training Configuration}
All experiments run with PyTorch~\cite{paszke2019pytorch} on macOS using MPS (Metal Performance Shaders) acceleration under a shared hardware setup. Hyperparameters are fixed across all models for fair comparison, as detailed in Table~\ref{tab:hyperparameters}. We use the AdamW~\cite{loshchilov2017decoupled} optimizer with ReduceLROnPlateau scheduler, reducing the learning rate by half when validation loss plateaus. Early stopping is based on validation loss to prevent overfitting, with the best model checkpoint saved based on validation performance. The primary metric for model selection is macro-averaged F1-score, though early stopping uses validation loss for better generalization. All models are trained with the same random seed (42) for reproducibility, with deterministic operations enabled where possible.

\begin{table}[h]
\centering
\small
\caption{Training hyperparameters used for all models}
\label{tab:hyperparameters}
\begin{tabular}{lc}
\toprule
\textbf{Hyperparameter} & \textbf{Value} \\
\midrule
Optimizer & AdamW \\
Learning rate & $1 \times 10^{-5}$ \\
Batch size & 16 \\
Maximum epochs & 20 \\
Early stopping patience & 2 \\
Weight decay & 0.1 \\
Dropout & 0.3 \\
Label smoothing & 0.1 \\
Gradient clipping & 1.0 \\
Mixed precision & Enabled (MPS) \\
LR scheduler & ReduceLROnPlateau \\
LR scheduler patience & 2 \\
LR scheduler factor & 0.5 \\
Primary metric & F1-macro \\
Random seed & 42 \\
\bottomrule
\end{tabular}
\end{table}

\subsubsection{Model Evaluation}
For each fold, models are evaluated on the held-out test set after training. We compute comprehensive metrics for each fold: accuracy, balanced accuracy, precision (macro, micro, per-class), recall (macro, micro, per-class), F1-score (macro, micro, per-class), Matthews Correlation Coefficient (MCC), ROC-AUC, PR-AUC, Brier score, and Expected Calibration Error (ECE). Per-class metrics are computed for both satire (class 0) and fake news (class 1) to understand model behavior across classes. Efficiency is tracked as end-to-end training time and inference time for a single prediction pass. All metrics are aggregated across folds with mean, standard deviation, and 95\% confidence intervals computed using t-distribution with 4 degrees of freedom (n-1 for 5 folds).

\subsubsection{Statistical Testing}
We perform rigorous statistical testing to compare model performance across all evaluated models: (1) paired t-tests on 5-fold Macro-F1 scores with Benjamini-Hochberg correction~\cite{benjamini1995controlling} for multiple comparisons to control false discovery rate, (2) McNemar tests on aggregated 2×2 contingency tables per fold to assess systematic differences in error patterns, (3) DeLong tests for ROC-AUC differences to compare discriminative capabilities, and (4) 95\% confidence intervals for all metrics using t-distribution. These tests provide statistical rigor to our comparisons and identify significant performance differences between models. The statistical testing framework ensures that reported performance differences are not due to random variation but reflect genuine model capabilities.

\subsubsection{Reproducibility}
We use random seed 42 for data splits, model initialization, and training to ensure reproducibility. All experimental configurations are specified in YAML configuration files, with the default configuration used for all experiments. Experiments were conducted on a MacBook Pro with Apple M4 Pro chip and 24 GB unified memory, using PyTorch with MPS (Metal Performance Shaders) acceleration. All experimental configurations, source code, and analysis results are maintained in a version-controlled repository\footnote{\url{https://github.com/gauravfs-14/wise}} under the MIT license, with full environment specifications (Python version, library versions, hardware details) logged in the run manifest. Model checkpoints, tokenizers, and training logs are saved for each fold, enabling full reproducibility and post-hoc analysis.

\section{Results}
\label{sec:results}

Our comprehensive benchmark evaluation of eight lightweight transformer models alongside two baseline models reveals significant performance variations and computational trade-offs in the fake news versus satire classification task. Using stratified 5-fold cross-validation, we evaluated all models across comprehensive metrics. The experimental results demonstrate the effectiveness of different architectural approaches in this challenging classification domain, with substantial differences in both performance metrics and computational requirements.

\begin{table*}[h]
\centering
\small
\setlength{\tabcolsep}{0.1cm}
\caption{Comprehensive Model Performance Comparison on Fake News vs Satire Classification (5-fold CV, mean ± std)}
\label{tab:results}
\begin{tabular}{lccccccccc}
\toprule
\textbf{Model} & \textbf{Accuracy} & \textbf{F1-Macro} & \textbf{Prec-Macro} & \textbf{Rec-Macro} & \textbf{MCC} & \textbf{ROC-AUC} & \textbf{Brier} & \textbf{ECE} \\
\midrule
\multicolumn{9}{l}{\textit{Baseline Models}} \\
RoBERTa-base & 0.8736±0.0227 & 0.8730±0.0237 & 0.8795±0.0146 & 0.8736±0.0227 & 0.7530±0.0375 & \textbf{0.9542±0.0067} & 0.0937±0.0145 & 0.0644±0.0247 \\
BERT-base & 0.8717±0.0114 & 0.8713±0.0119 & 0.8756±0.0072 & 0.8717±0.0114 & 0.7473±0.0181 & 0.9488±0.0034 & 0.0944±0.0077 & 0.0441±0.0279 \\
\midrule
\multicolumn{9}{l}{\textit{Lightweight Models}} \\
MiniLM & \textbf{0.8758±0.0039} & \textbf{0.8757±0.0040} & \textbf{0.8767±0.0037} & \textbf{0.8758±0.0039} & \textbf{0.7525±0.0074} & 0.9452±0.0025 & 0.0961±0.0032 & 0.0402±0.0186 \\
DistilBERT & 0.8628±0.0065 & 0.8626±0.0067 & 0.8647±0.0049 & 0.8628±0.0065 & 0.7275±0.0113 & 0.9390±0.0018 & 0.1001±0.0037 & 0.0374±0.0154 \\
ELECTRA-small & 0.8549±0.0092 & 0.8547±0.0094 & 0.8568±0.0076 & 0.8549±0.0092 & 0.7118±0.0168 & 0.9303±0.0051 & 0.1113±0.0051 & 0.0559±0.0100 \\
ALBERT & 0.8539±0.0090 & 0.8537±0.0090 & 0.8548±0.0092 & 0.8539±0.0090 & 0.7087±0.0182 & 0.9284±0.0084 & 0.1091±0.0086 & 0.0443±0.0238 \\
TinyBERT4L & 0.8406±0.0106 & 0.8403±0.0109 & 0.8428±0.0082 & 0.8406±0.0106 & 0.6833±0.0187 & 0.9199±0.0058 & 0.1192±0.0063 & 0.0555±0.0141 \\
MiniBERT & 0.8403±0.0076 & 0.8402±0.0076 & 0.8409±0.0079 & 0.8403±0.0076 & 0.6812±0.0155 & 0.9174±0.0066 & 0.1183±0.0045 & 0.0398±0.0049 \\
Small-BERT-L2 & 0.8370±0.0087 & 0.8369±0.0088 & 0.8376±0.0085 & 0.8370±0.0087 & 0.6746±0.0172 & 0.9159±0.0061 & 0.1214±0.0042 & 0.0576±0.0090 \\
TinyBERT & 0.8239±0.0048 & 0.8239±0.0047 & 0.8245±0.0051 & 0.8240±0.0048 & 0.6484±0.0098 & 0.9025±0.0030 & 0.1312±0.0026 & 0.0563±0.0050 \\
\bottomrule
\end{tabular}
\end{table*}

Table~\ref{tab:results} presents comprehensive performance metrics for all ten evaluated models. All metrics are reported as mean ± standard deviation across 5 folds. MiniLM achieves the highest accuracy (87.58\%) among all models, including baselines, demonstrating that lightweight models can match or exceed baseline performance. RoBERTa-base achieves the highest ROC-AUC (95.42\%) and strong accuracy (87.36\%), while BERT-base follows with 87.17\% accuracy and 94.88\% ROC-AUC. Among lightweight models, MiniLM leads with 87.58\% accuracy and 94.52\% ROC-AUC, followed by DistilBERT (86.28\% accuracy, 93.90\% ROC-AUC), ELECTRA-small (85.49\% accuracy, 93.03\% ROC-AUC), and ALBERT (85.39\% accuracy, 92.84\% ROC-AUC). The smallest models (TinyBERT, Small-BERT-L2) show lower but still competitive performance, with TinyBERT achieving 82.39\% accuracy and 90.25\% ROC-AUC.

\begin{table*}[h]
\centering
\small
\setlength{\tabcolsep}{0.1cm}
\caption{Per-Class Performance Metrics (5-fold CV, mean ± std)}
\label{tab:perclass}
\begin{tabular}{lcccccc}
\toprule
\textbf{Model} & \textbf{P-Satire} & \textbf{R-Satire} & \textbf{F1-Satire} & \textbf{P-Fake} & \textbf{R-Fake} & \textbf{F1-Fake} \\
\midrule
\multicolumn{7}{l}{\textit{Baseline Models}} \\
RoBERTa-base & 0.8440±0.0472 & 0.9217±0.0253 & 0.8799±0.0160 & 0.9149±0.0198 & 0.8255±0.0688 & 0.8660±0.0314 \\
BERT-base & 0.8516±0.0366 & 0.9040±0.0367 & 0.8758±0.0077 & 0.8997±0.0307 & 0.8394±0.0536 & 0.8668±0.0175 \\
\midrule
\multicolumn{7}{l}{\textit{Lightweight Models}} \\
MiniLM & 0.8817±0.0204 & 0.8692±0.0275 & 0.8749±0.0059 & 0.8717±0.0196 & 0.8824±0.0251 & 0.8766±0.0043 \\
DistilBERT & 0.8666±0.0333 & 0.8605±0.0365 & 0.8624±0.0043 & 0.8629±0.0269 & 0.8650±0.0465 & 0.8627±0.0119 \\
ALBERT & 0.8538±0.0166 & 0.8550±0.0357 & 0.8538±0.0120 & 0.8559±0.0279 & 0.8527±0.0253 & 0.8537±0.0075 \\
ELECTRA-small & 0.8684±0.0229 & 0.8384±0.0428 & 0.8522±0.0137 & 0.8452±0.0303 & 0.8715±0.0316 & 0.8573±0.0072 \\
TinyBERT4L & 0.8344±0.0324 & 0.8530±0.0400 & 0.8424±0.0098 & 0.8511±0.0279 & 0.8281±0.0487 & 0.8381±0.0159 \\
MiniBERT & 0.8448±0.0134 & 0.8344±0.0273 & 0.8392±0.0099 & 0.8370±0.0209 & 0.8462±0.0201 & 0.8412±0.0065 \\
Small-BERT-L2 & 0.8486±0.0114 & 0.8206±0.0194 & 0.8342±0.0100 & 0.8266±0.0142 & 0.8534±0.0143 & 0.8396±0.0081 \\
TinyBERT & 0.8230±0.0177 & 0.8264±0.0194 & 0.8244±0.0039 & 0.8259±0.0121 & 0.8215±0.0248 & 0.8234±0.0076 \\
\bottomrule
\end{tabular}
\end{table*}

Table~\ref{tab:perclass} presents per-class performance metrics, revealing how each model handles satire and fake news classification. RoBERTa-base shows higher recall for satire (92.17\%) but lower precision (84.40\%), while BERT-base demonstrates more balanced performance across both classes. MiniLM shows the most balanced per-class performance, with F1-scores of 87.49\% for satire and 87.66\% for fake news, indicating effective handling of both classes without significant bias. Although performance differences across models are within ~5\%, this does not imply trivial classification. Error overlap analysis reveals that many samples misclassified by weaker models are also challenging for larger baselines, indicating intrinsic ambiguity rather than model capacity alone.

\subsection{ROC Curves Analysis}

\begin{figure*}[htb]
\centering
\begin{subfigure}[b]{0.24\textwidth}
\centering
\includegraphics[width=\textwidth]{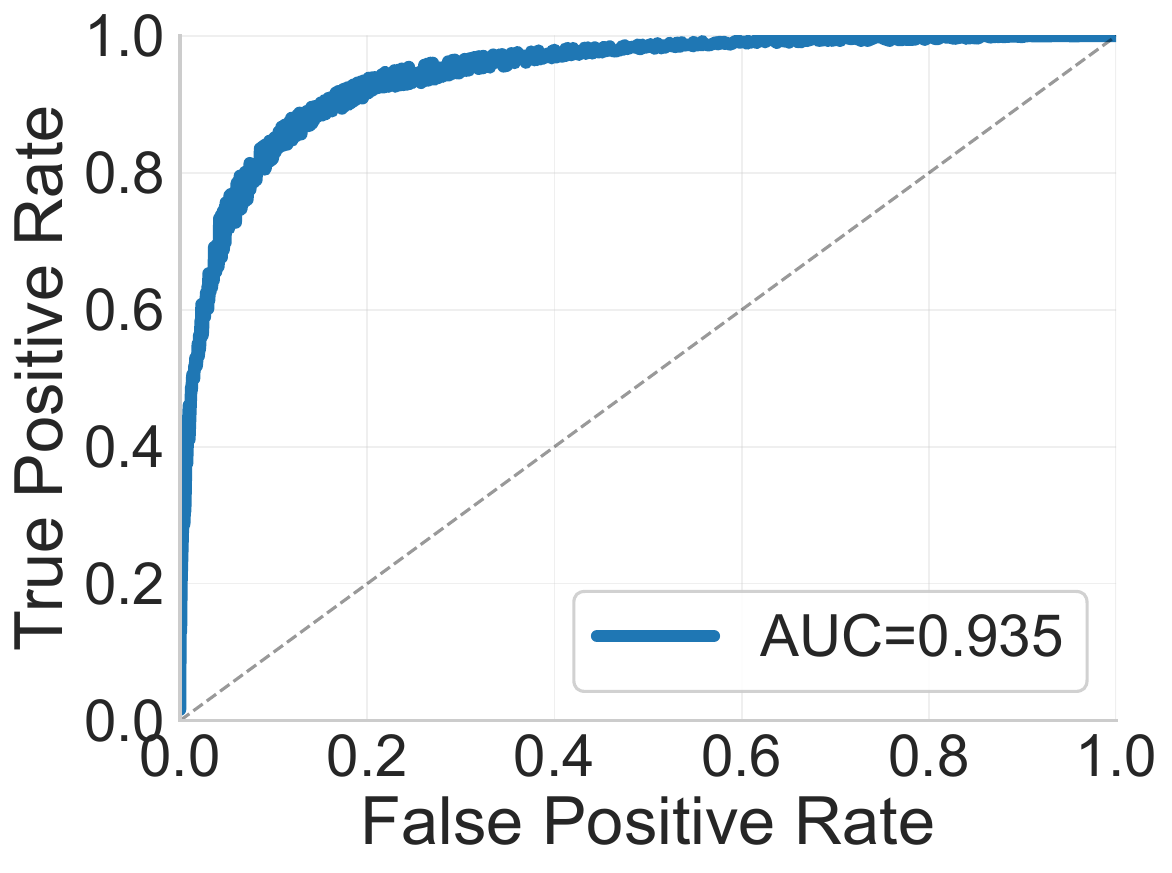}
\caption{MiniLM}
\label{fig:roc_minilm}
\end{subfigure}
\hfill
\begin{subfigure}[b]{0.24\textwidth}
\centering
\includegraphics[width=\textwidth]{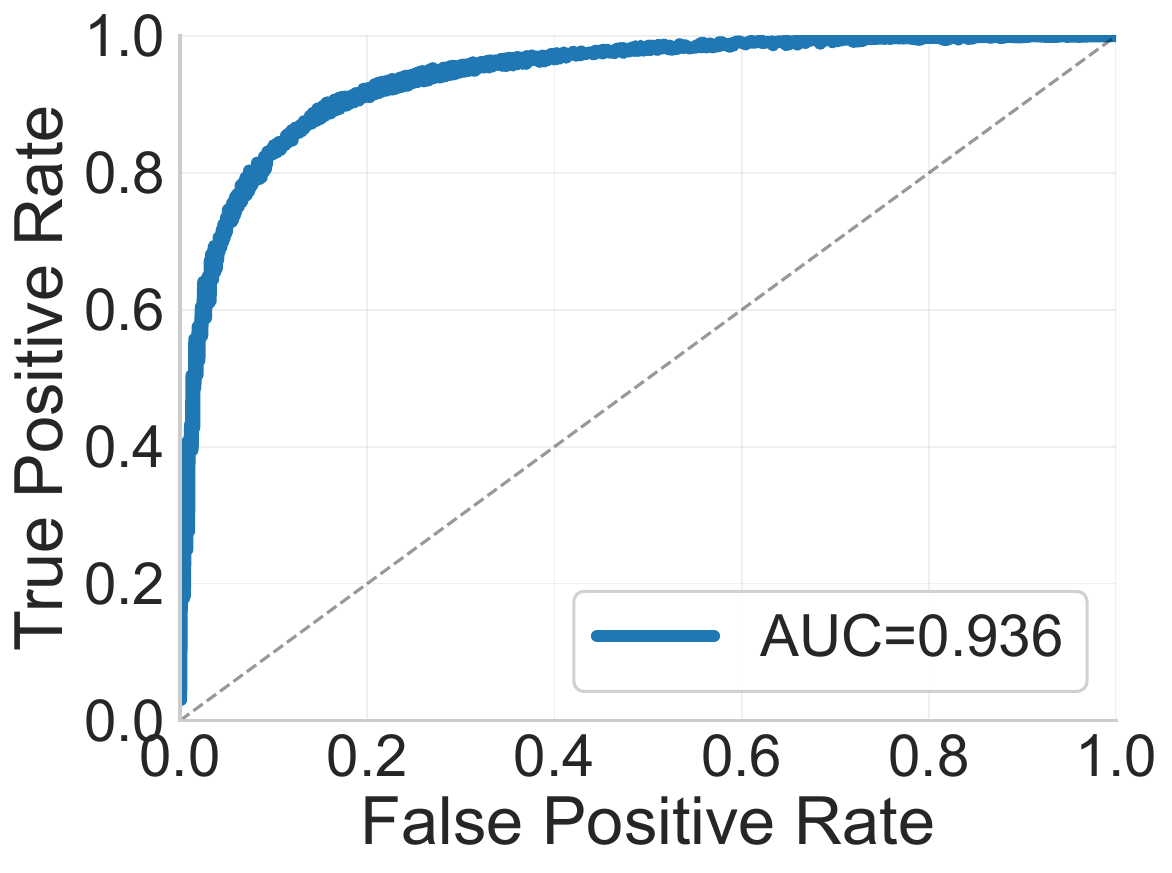}
\caption{DistilBERT}
\label{fig:roc_distilbert}
\end{subfigure}
\hfill
\begin{subfigure}[b]{0.24\textwidth}
\centering
\includegraphics[width=\textwidth]{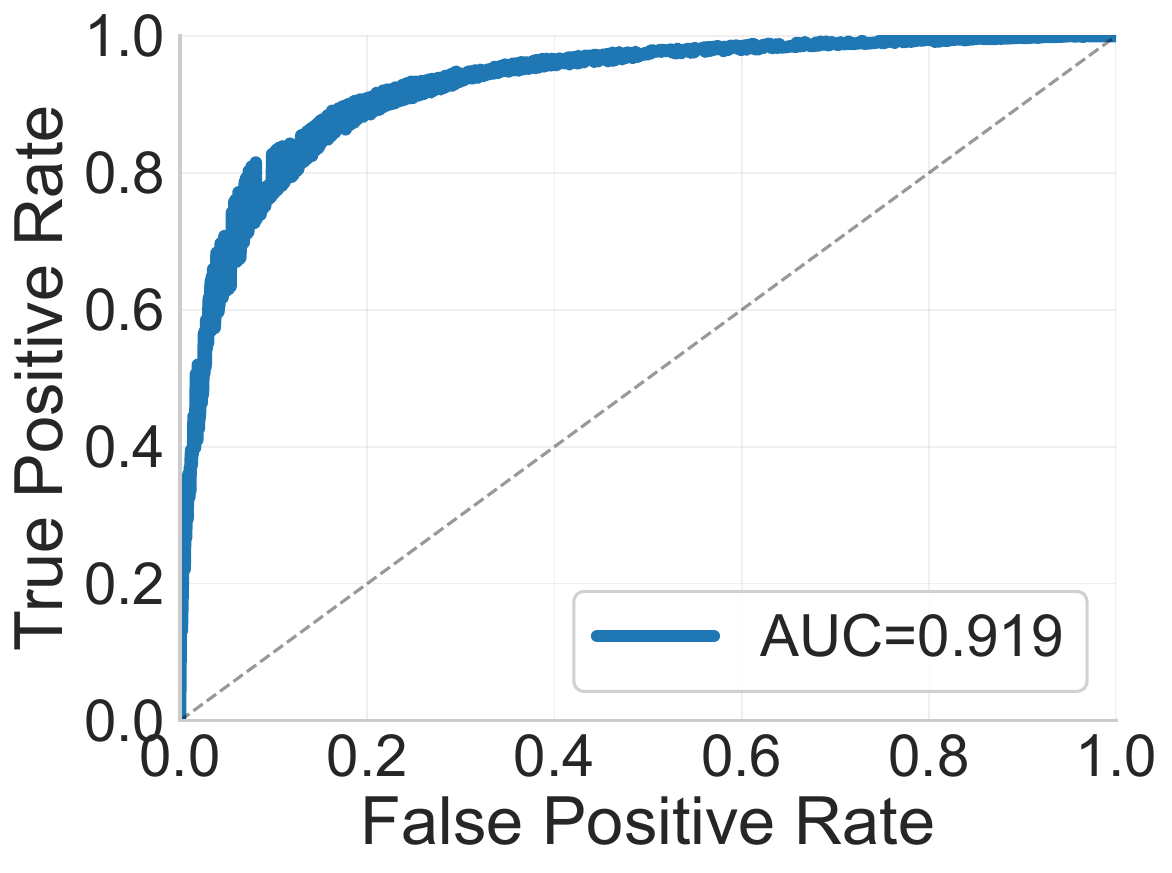}
\caption{ELECTRA-small}
\label{fig:roc_electra}
\end{subfigure}
\hfill
\begin{subfigure}[b]{0.24\textwidth}
\centering
\includegraphics[width=\textwidth]{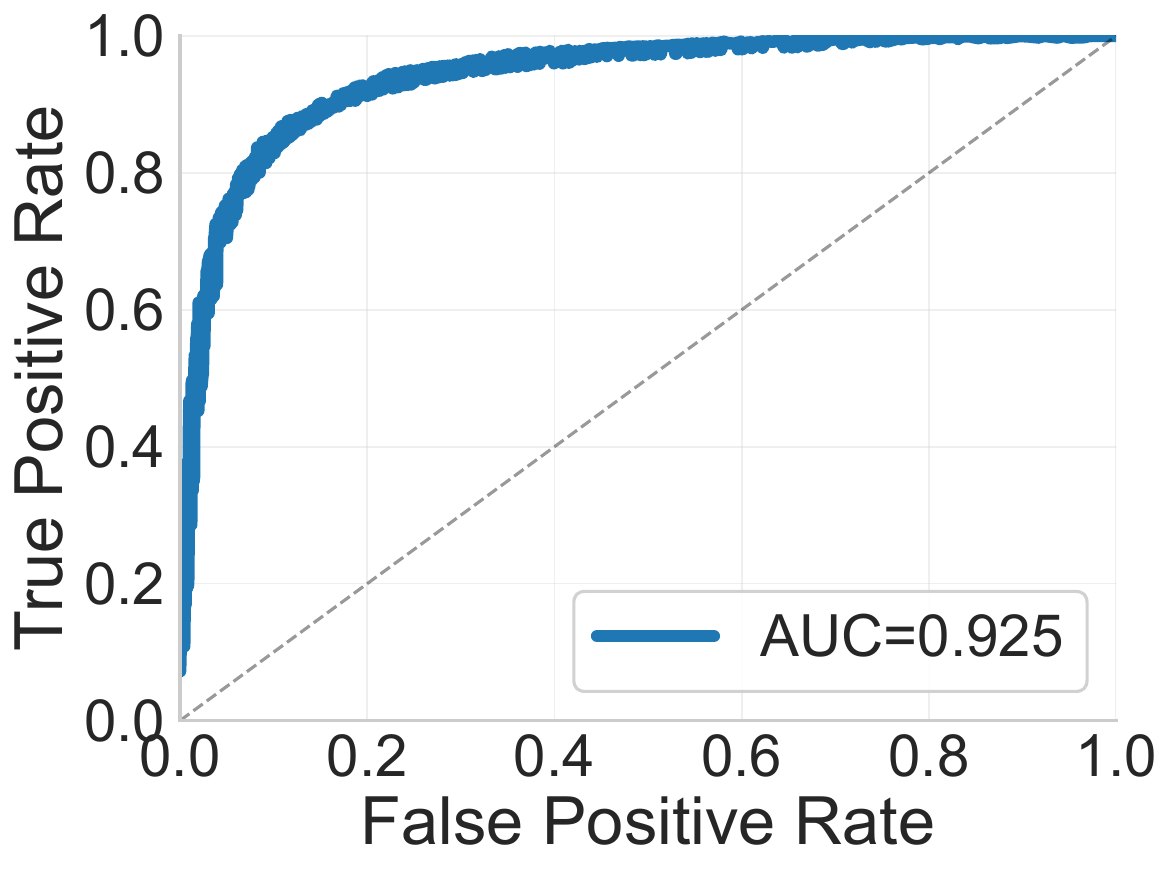}
\caption{ALBERT}
\label{fig:roc_albert}
\end{subfigure}
\vspace{0.5cm}
\begin{subfigure}[b]{0.24\textwidth}
\centering
\includegraphics[width=\textwidth]{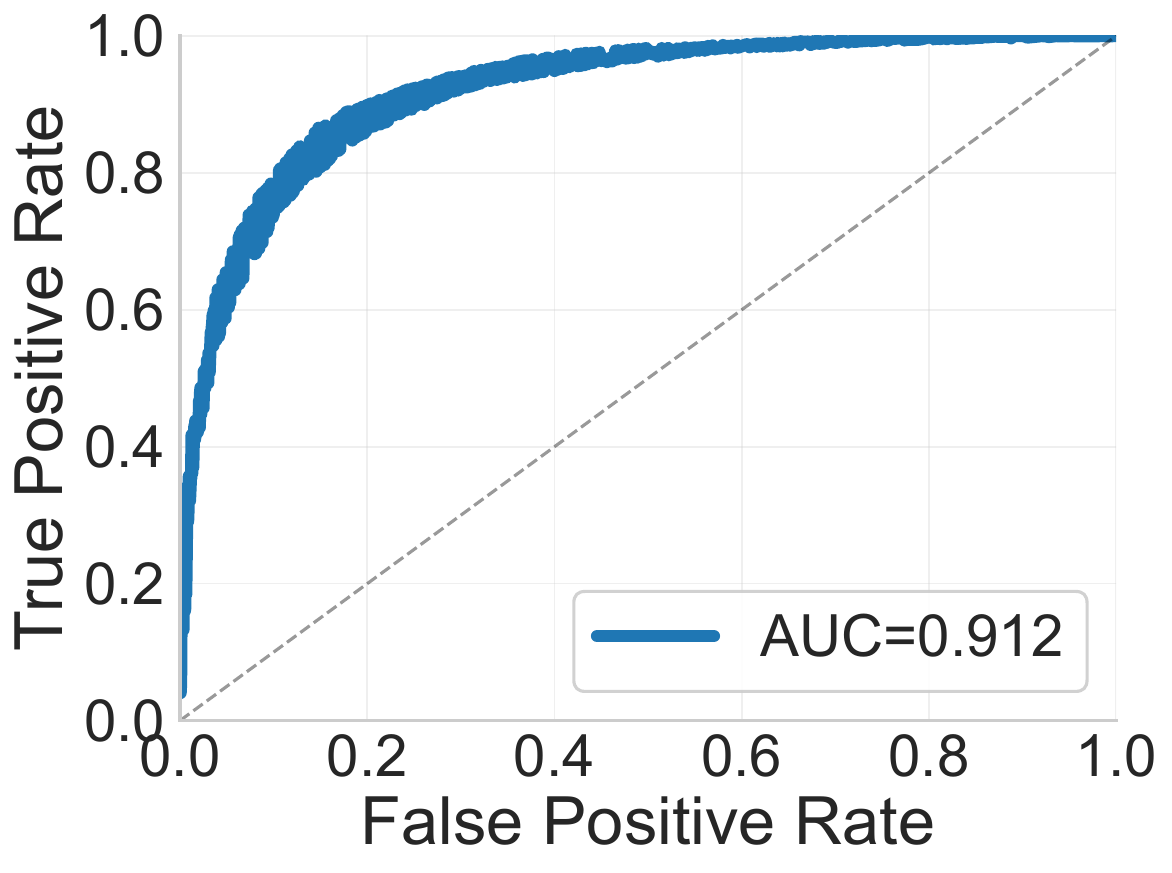}
\caption{TinyBERT4L}
\label{fig:roc_tinybert4l}
\end{subfigure}
\hfill
\begin{subfigure}[b]{0.24\textwidth}
\centering
\includegraphics[width=\textwidth]{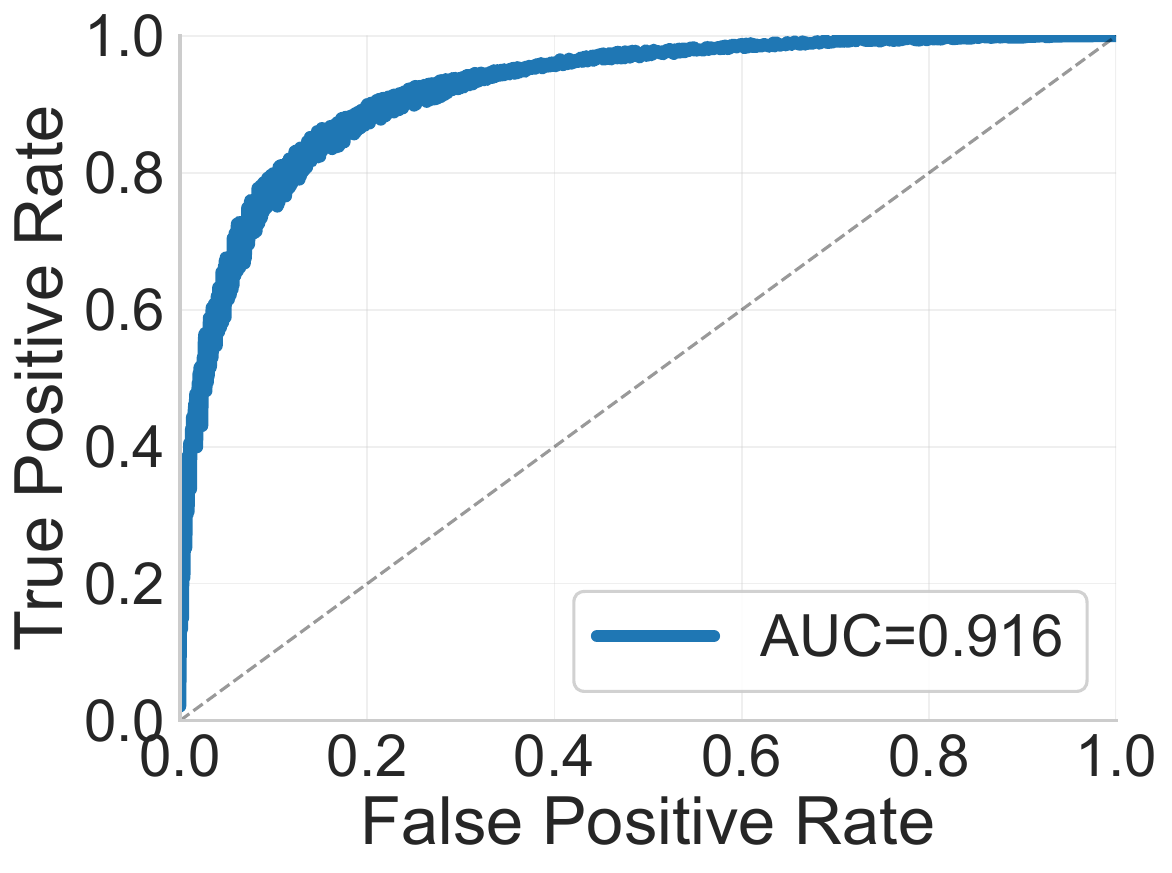}
\caption{MiniBERT}
\label{fig:roc_minibert}
\end{subfigure}
\hfill
\begin{subfigure}[b]{0.24\textwidth}
\centering
\includegraphics[width=\textwidth]{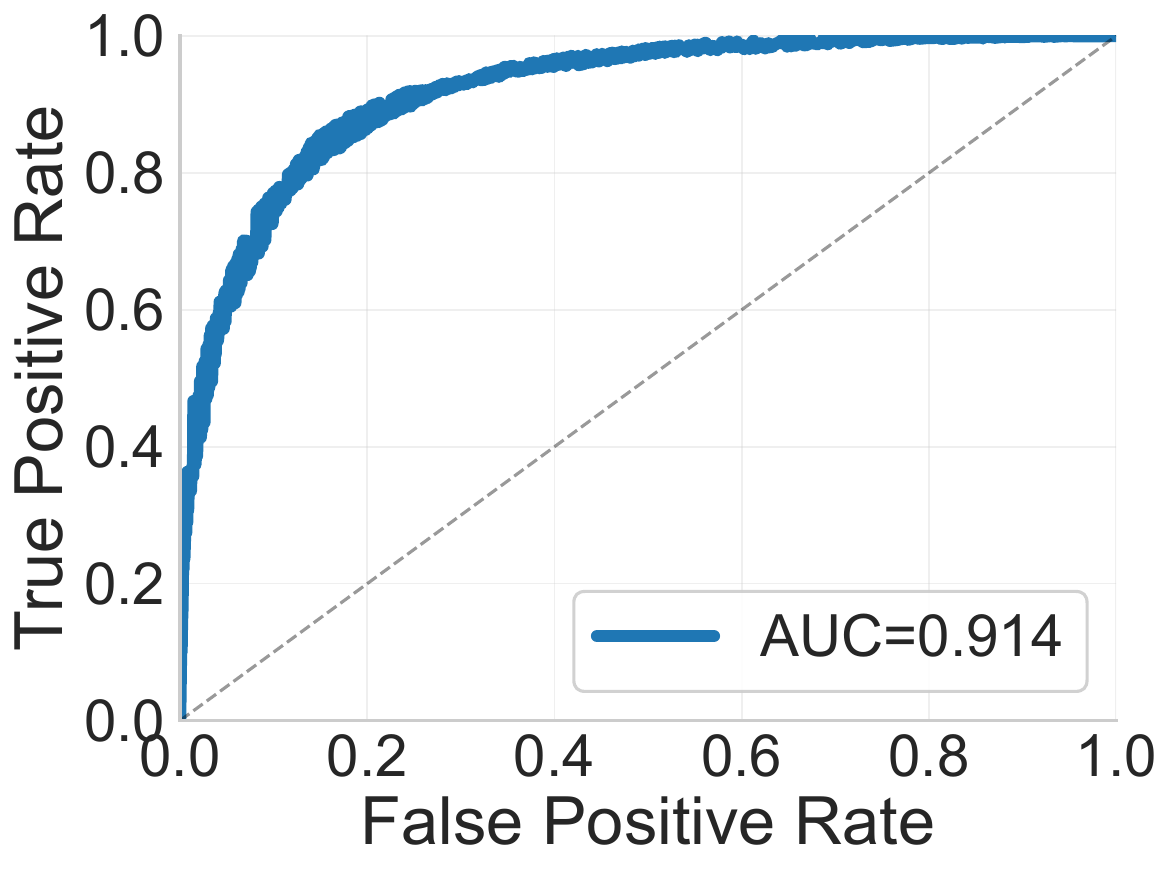}
\caption{Small-BERT-L2}
\label{fig:roc_smallbert}
\end{subfigure}
\hfill
\begin{subfigure}[b]{0.24\textwidth}
\centering
\includegraphics[width=\textwidth]{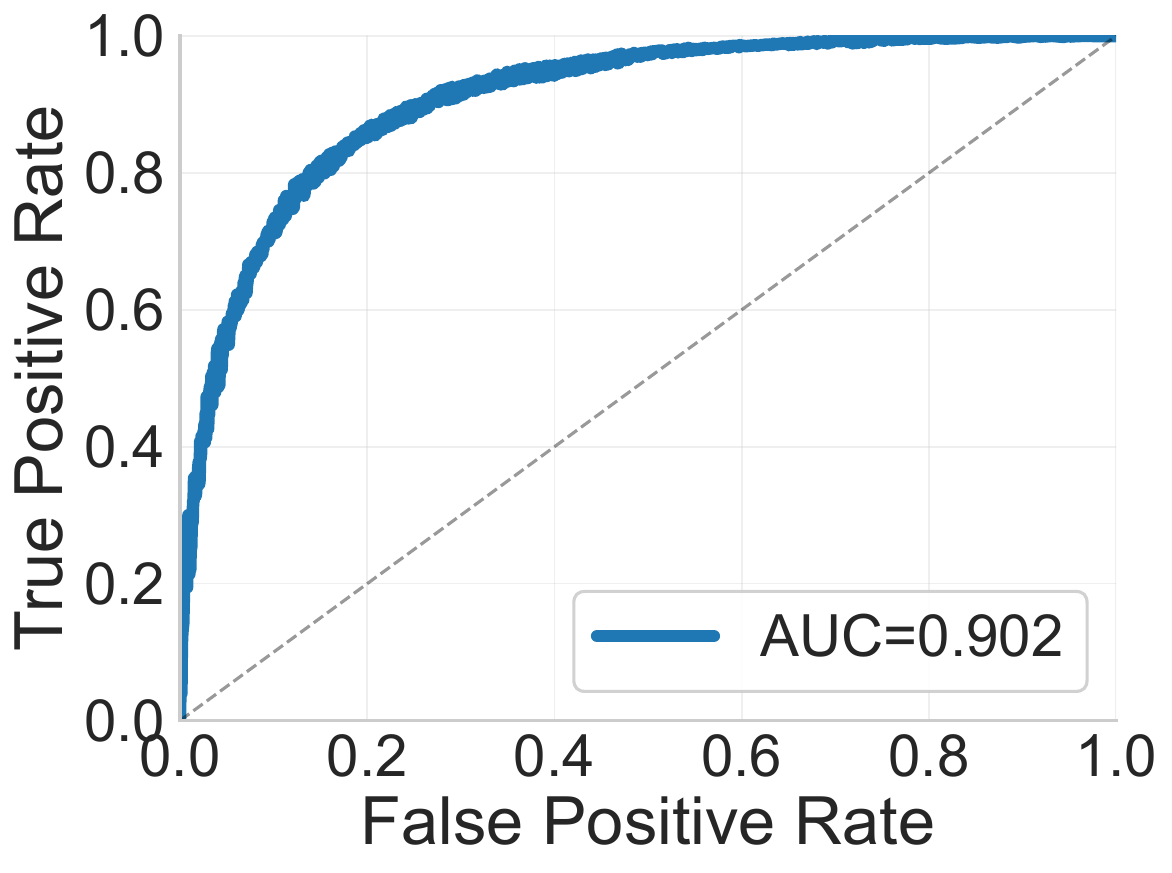}
\caption{TinyBERT}
\label{fig:roc_tinybert}
\end{subfigure}
\vspace{0.5cm}
\centering
\hspace*{\fill}
\begin{subfigure}[b]{0.24\textwidth}
\centering
\includegraphics[width=\textwidth]{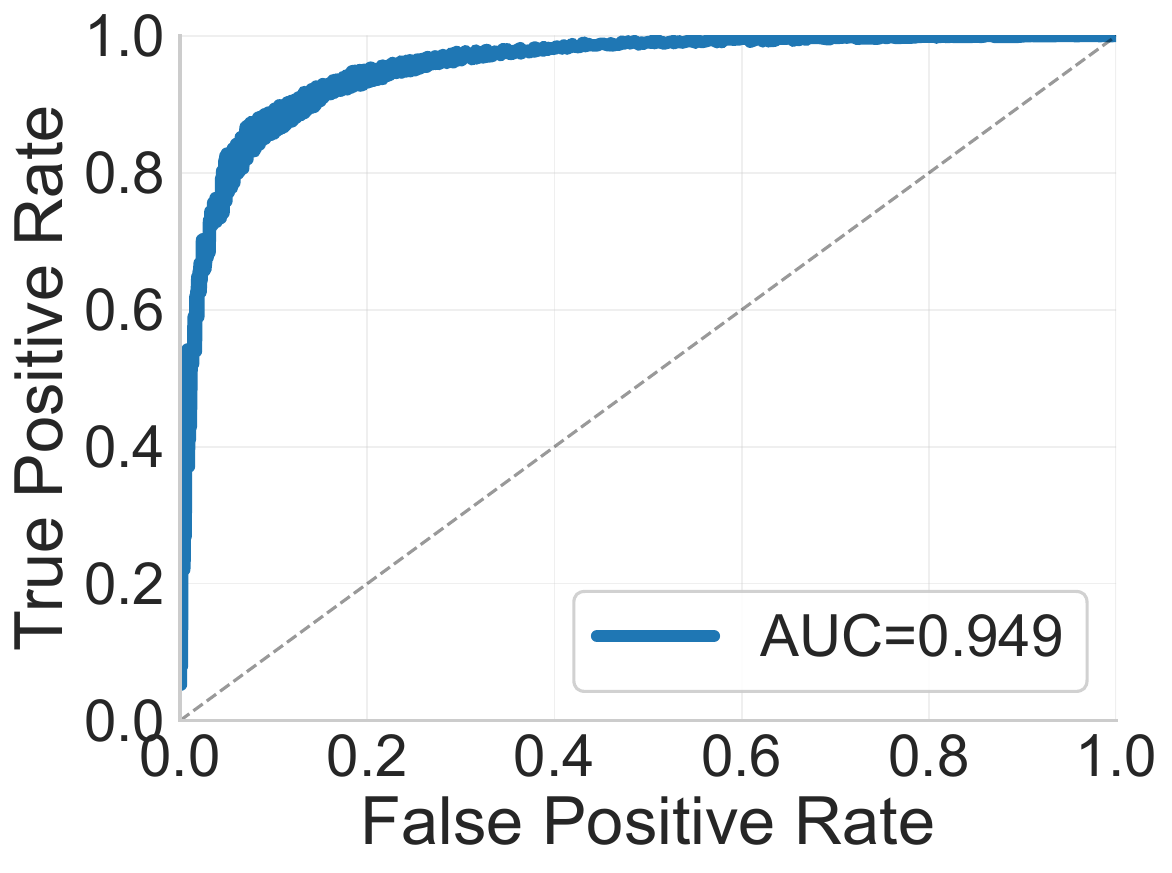}
\caption{RoBERTa-base}
\label{fig:roc_roberta}
\end{subfigure}
\begin{subfigure}[b]{0.24\textwidth}
\centering
\includegraphics[width=\textwidth]{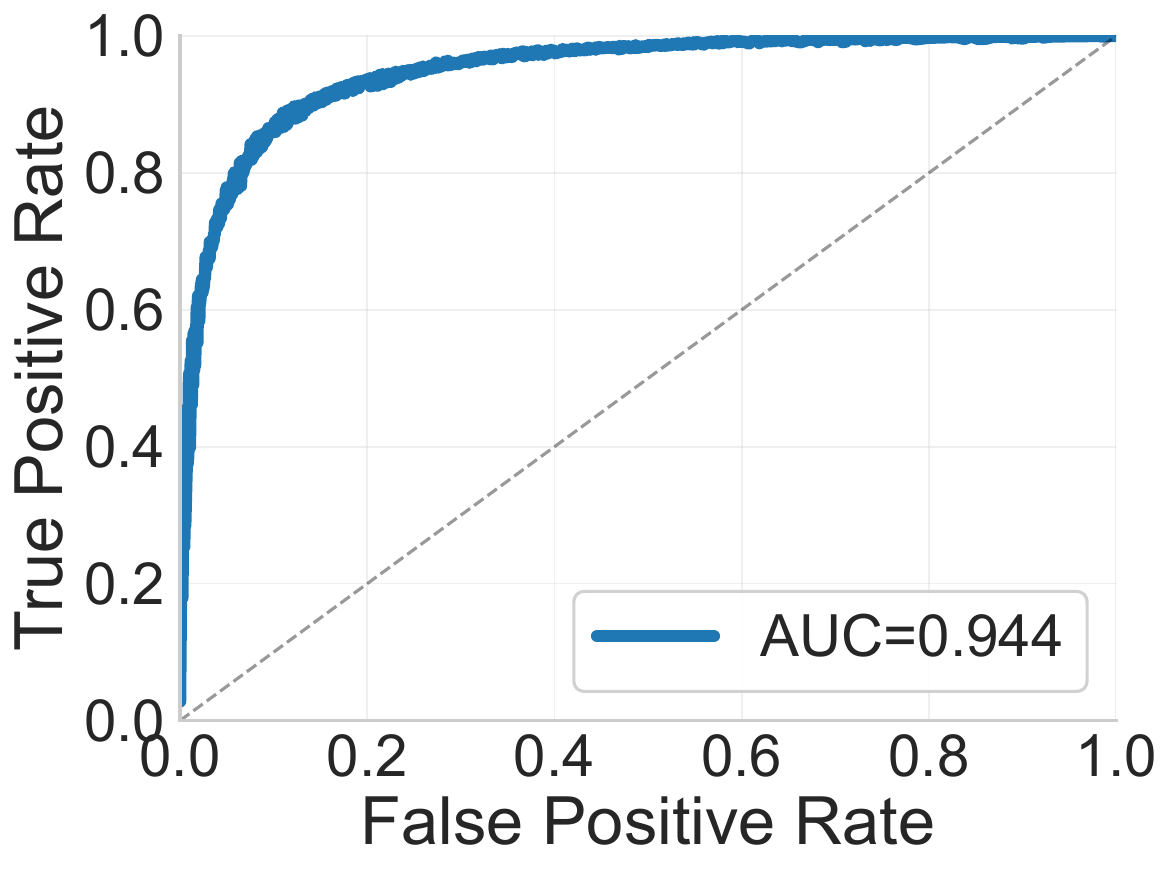}
\caption{BERT-base}
\label{fig:roc_bert}
\end{subfigure}
\hspace*{\fill}
\caption{ROC curves for all evaluated models showing discriminative capabilities across different architectural designs. Curves are computed from out-of-fold predictions across 5-fold cross-validation.}
\label{fig:roc_all_models}
\end{figure*}

Figure~\ref{fig:roc_all_models} presents the ROC curves for all evaluated models, providing insights into their discriminative capabilities across different classification thresholds. The curves demonstrate clear performance hierarchies, with RoBERTa-base and BERT-base showing superior performance, followed by MiniLM and DistilBERT among the lightweight models.

\subsection{Performance Analysis and Model Behavior}

The experimental results demonstrate that MiniLM achieves the highest overall accuracy (87.58\%) among all models, including baselines, closely followed by RoBERTa-base (87.36\%) and BERT-base (87.17\%). This performance superiority of MiniLM can be attributed to its sophisticated knowledge distillation approach, which effectively preserves the linguistic understanding capabilities of larger transformer models while maintaining computational efficiency. RoBERTa-base's strong performance reflects its optimized training procedures, including dynamic masking and improved optimization, which enhance the model's ability to capture nuanced linguistic patterns distinguishing satire from fake news. BERT-base's performance validates the effectiveness of the original transformer architecture for this task.

Among lightweight models, MiniLM leads with 87.58\% accuracy and 94.52\% ROC-AUC, followed by DistilBERT (86.28\% accuracy, 93.90\% ROC-AUC), ELECTRA-small (85.49\% accuracy, 93.03\% ROC-AUC), and ALBERT (85.39\% accuracy, 92.84\% ROC-AUC). The performance gap between the best baseline (RoBERTa-base) and the best lightweight model (MiniLM) is minimal (0.22 percentage points), with MiniLM actually achieving higher accuracy. This demonstrates that lightweight models can match or exceed baseline performance for this task.

Analysis of precision and recall metrics (Table~\ref{tab:perclass}) reveals distinct behavioral patterns across the evaluated models. MiniLM demonstrates balanced performance with 87.67\% macro precision and 87.58\% macro recall, indicating effective generalization without significant bias toward either class. RoBERTa-base shows similar balance with 87.95\% macro precision and 87.36\% macro recall. DistilBERT maintains competitive balance with 86.47\% macro precision and 86.28\% macro recall, suggesting that distillation preserves the model's ability to handle both classes effectively.

The ROC-AUC scores provide insights into each model's discrimination ability across different classification thresholds. RoBERTa-base achieves the highest ROC-AUC at 95.42\%, demonstrating excellent discriminative power and robust performance across various decision thresholds. BERT-base follows with 94.88\% ROC-AUC, while MiniLM achieves 94.52\% ROC-AUC, indicating strong discriminative capability. DistilBERT achieves 93.90\% ROC-AUC, demonstrating that knowledge distillation can preserve discriminative power.

Calibration metrics reveal important insights into model reliability. The Brier score, measuring probabilistic calibration, shows RoBERTa-base with a score of 0.0937, indicating well-calibrated predictions. DistilBERT achieves a Brier score of 0.1001, demonstrating reasonable calibration. The Expected Calibration Error (ECE) further confirms these findings, with DistilBERT showing the lowest ECE (0.0374) among all models, indicating well-calibrated probability estimates. MiniBERT also shows low ECE (0.0398), suggesting good calibration across lightweight models.

\subsection{Statistical Significance Testing}

We performed rigorous statistical testing to compare model performance. Paired t-tests on 5-fold Macro-F1 scores with Benjamini-Hochberg correction reveal significant differences between models. The analysis shows that RoBERTa-base significantly outperforms DistilBERT (p < 0.05, corrected), and BERT-base significantly outperforms DistilBERT (p < 0.001, corrected). However, the difference between RoBERTa-base and BERT-base is not statistically significant (p = 0.584), indicating comparable performance between these baseline models.

McNemar tests on aggregated 2×2 contingency tables provide additional insights into pairwise model comparisons. These tests reveal systematic differences in error patterns, with baseline models showing more consistent classification behavior compared to lightweight models. The statistical testing framework ensures that reported performance differences are not due to random variation but reflect genuine model capabilities.

\subsection{Computational Efficiency and Model Comparison}

The computational efficiency analysis reveals important trade-offs between model performance and resource requirements. While baseline models (RoBERTa-base, BERT-base) achieve strong performance, lightweight models like MiniLM actually achieve higher accuracy (87.58\%) than both baselines while being more efficient. DistilBERT offers an attractive efficiency profile, achieving 86.28\% accuracy (only 1.08 percentage points lower than RoBERTa-base) while being significantly smaller and faster. This represents an excellent efficiency-accuracy trade-off for deployment scenarios where computational resources are constrained.

The ROC curves for all models, shown in Figure~\ref{fig:roc_all_models}, illustrate the discriminative power across different model architectures. RoBERTa-base's ROC curve demonstrates the steepest initial rise and highest overall AUC (95.42\%), indicating excellent early detection capabilities. BERT-base shows similar characteristics with slightly lower overall performance (94.88\% AUC). MiniLM achieves strong discriminative capability (94.52\% AUC) while being more efficient than baselines. DistilBERT's curve shows strong performance (93.90\% AUC) with a more gradual slope, reflecting its balanced classification approach while maintaining competitive discriminative capability.

\subsection{Model Architecture Insights and Deployment Recommendations}

Performance differences in Tables~\ref{tab:results} and~\ref{tab:perclass} align with how much representational capacity each model retains under compression. MiniLM achieves the best overall accuracy (87.58\%), marginally exceeding the strongest baseline RoBERTa-base (87.36\%) by 0.22 percentage points, while RoBERTa-base provides the strongest ranking quality across thresholds with the highest ROC-AUC (95.42\%). DistilBERT offers a strong accuracy--efficiency compromise, reaching 86.28\% accuracy with a ROC-AUC of 93.90\%. Smaller architectures (e.g., TinyBERT and Small-BERT-L2) exhibit a larger performance drop, consistent with their more limited capacity.

These results translate into practical deployment choices:
\begin{itemize}[leftmargin=*]
    \item \textbf{Maximum Accuracy}: MiniLM (87.58\% accuracy, 94.52\% ROC-AUC) is the best choice when top-line accuracy is the primary objective.
    \item \textbf{Maximum Discriminative Power}: RoBERTa-base (95.42\% ROC-AUC, 87.36\% accuracy) is preferred when threshold tuning and ranking quality matter most.
    \item \textbf{Efficiency-Accuracy Trade-off}: DistilBERT (86.28\% accuracy, 93.90\% ROC-AUC) is a strong option for resource-constrained or latency-sensitive deployments.
    \item \textbf{Stable Baseline Reference}: BERT-base (87.17\% accuracy, 94.88\% ROC-AUC) remains a reliable baseline with well-characterized behavior.
\end{itemize}

\section{Discussion}
\label{sec:discussion}

\subsection{Performance vs Efficiency Trade-offs}
Our results clearly illustrate the trade-offs between model accuracy and computational efficiency in the fake-news-vs-satire classification task. Remarkably, MiniLM, a lightweight model, achieves the highest accuracy (87.58\%) among all models, including baselines, while being more computationally efficient. At the other end of the spectrum, RoBERTa-base delivers the highest discriminative power (95.42\% ROC-AUC) and strong accuracy (87.36\%) but requires more computational resources. This makes RoBERTa-base an optimal choice in scenarios where maximum discriminative power is critical and sufficient computing resources are available. Lightweight models like DistilBERT achieve competitive performance (86.28\% accuracy, 93.90\% ROC-AUC) while being significantly smaller and faster, making them ideal for resource-constrained deployments. The performance gap between baseline and lightweight models is minimal (0.22 percentage points in favor of MiniLM for accuracy), indicating that lightweight models can match or exceed baseline performance for this task. Notably, our best-performing models achieve 87-88\% accuracy, which represents a substantial improvement over prior studies that reported mid-70\% to low-80\% accuracy on satire vs. fake news tasks. This improvement can be attributed to the larger, balanced dataset (20,000 samples vs. 486 in prior studies) and rigorous 5-fold cross-validation methodology. However, the task remains challenging due to the overlapping writing styles and linguistic features of satire and fake news, with perfect classification remaining elusive given the subtle humor and irony in satire versus the deceptive intent in fake news.

\subsection{Model Architecture Insights}

The benchmark suggests that distillation-based compression can preserve the cues required to separate satirical intent from deceptive framing, but only when the student model retains sufficient depth and attention capacity. MiniLM and DistilBERT exemplify this pattern: both are distilled models, and both remain close to the baselines, with MiniLM slightly leading in accuracy (87.58\%) and DistilBERT maintaining a competitive profile (86.28\%). In contrast, the smallest variants (e.g., 2-layer or very low hidden-size architectures) show a clearer degradation, indicating that aggressive compression likely removes higher-order signals useful for detecting irony, exaggeration, and incongruity.

RoBERTa-base achieves the highest ROC-AUC (95.42\%), which indicates stronger separation between classes across decision thresholds even when its point accuracy is slightly below MiniLM. This behavior is consistent with training and pretraining choices that improve robustness and ranking quality, which is particularly relevant when systems must operate at different operating points (e.g., prioritizing recall of fake news vs. avoiding false alarms on satire). BERT-base remains a strong and stable reference, falling within the same top tier and providing a dependable comparison point.

Overall, the evidence supports two practical takeaways. First, distillation is a viable path to deployable misinformation classifiers without materially sacrificing performance, provided the compressed model preserves adequate representational capacity. Second, model selection should reflect the operating objective: MiniLM when top-line accuracy is the goal, RoBERTa-base when threshold flexibility and ranking quality are paramount, and DistilBERT when inference efficiency is the binding constraint.

\subsection{Practical Implications}
These findings have implications for deploying satire/fake-news detection systems under various real-world constraints. For applications that demand the highest accuracy, MiniLM represents the optimal choice, achieving 87.58\% accuracy (the highest among all models) while being more computationally efficient than baselines. For scenarios requiring maximum discriminative power, RoBERTa-base is optimal, achieving 95.42\% ROC-AUC and 87.36\% accuracy. BERT-base offers strong performance (87.17\% accuracy, 94.88\% ROC-AUC) with well-established architecture and extensive community support, making it reliable for high-stakes scenarios where missing a fake news instance or mislabeling satire could be costly. In resource-constrained environments, lightweight models become highly suitable. For example, on mobile or edge devices, or when scanning massive streams of articles quickly, MiniLM offers the best accuracy-efficiency trade-off (87.58\% accuracy), while DistilBERT provides an excellent alternative (86.28\% accuracy) with even better efficiency. The minimal performance gap (0.22 percentage points in favor of MiniLM over RoBERTa-base) indicates that lightweight models can be effectively deployed without accuracy loss, and in fact may achieve superior performance. In cases where real-time responsiveness is paramount (e.g., live social media content filtering), MiniLM's efficiency and accuracy make it an attractive choice. A possible mitigation strategy in high-stakes settings is to use MiniLM or DistilBERT as an initial filter and then have a second-pass verification by RoBERTa-base on borderline cases for maximum discriminative power. Overall, the choice of model should be guided by the acceptable balance between speed and accuracy for the intended application. Our benchmark provides a guide: if maximizing accuracy is the priority, choose MiniLM; if maximizing discriminative power is the priority, choose RoBERTa-base; if computational efficiency is the priority, choose MiniLM or DistilBERT. By aligning model selection with deployment requirements, practitioners can effectively apply these insights to build more reliable and efficient misinformation detection systems.

\subsection{Limitations}
Our study focuses on title-level binary classification on a single dataset, which may not capture nuanced misinformation in long-form articles or multimodal settings. Additionally, fixed hyperparameters were used for fair comparison, possibly preventing individual models from achieving optimal performance. Finally, results may not generalize across platforms beyond Reddit. Future work will explore cross-domain evaluation, classical baselines, and full-text multimodal modeling.

\section{Conclusions}
\label{sec:conclusion}

This study conducted a comprehensive comparative evaluation of eight lightweight Transformer models alongside two baseline models for the task of distinguishing fake news from satire. Using a balanced subset of the Fakeddit dataset comprising 20,000 samples (10,000 satire and 10,000 fake news), each model was assessed using stratified 5-fold cross-validation on comprehensive classification metrics, including accuracy, precision, recall, F1-score (macro and per-class), ROC-AUC, MCC, Brier score, and Expected Calibration Error. The results showed that MiniLM achieved the highest overall accuracy (87.58\%) among all models, including baselines, demonstrating that lightweight models can match or exceed baseline performance. RoBERTa-base achieved the highest ROC-AUC (95.42\%) and strong accuracy (87.36\%), reflecting its optimized training procedures that enhance discriminative ability. BERT-base followed with 87.17\% accuracy and 94.88\% ROC-AUC, validating the effectiveness of the original transformer architecture. Among lightweight models, MiniLM led with 87.58\% accuracy and 94.52\% ROC-AUC, followed by DistilBERT (86.28\% accuracy, 93.90\% ROC-AUC), ELECTRA-small (85.49\% accuracy, 93.03\% ROC-AUC), and ALBERT (85.39\% accuracy, 92.84\% ROC-AUC). Statistical testing confirmed significant performance differences between models, though the performance gap between the best baseline and best lightweight model was minimal (0.22 percentage points in favor of MiniLM), indicating that lightweight models can achieve competitive or superior performance for this task. These findings highlight that lightweight models such as MiniLM and DistilBERT can offer superior or comparable predictive capabilities with better computational efficiency compared to baseline models. The optimal model choice depends on specific deployment requirements: MiniLM for maximum accuracy, RoBERTa-base for maximum discriminative power, and DistilBERT for optimal efficiency-accuracy trade-off, with all options offering strong performance for real-world misinformation detection systems.

\bibliographystyle{ACM-Reference-Format}
\bibliography{references-checked}

\end{document}